# Manifold regularized kernel logistic regression for web image annotation

W. Liu[1], H. Liu[1], D.Tao[2*], Y. Wang[1], K. Lu[3]

[1]China University of Petroleum (East China)

**Abstract**

With the rapid advance of Internet technology and smart devices, users often need to manage large amounts of multimedia information using smart devices, such as personal image and video accessing and browsing. These requirements heavily rely on the success of image (video) annotation, and thus large scale image annotation through innovative machine learning methods has attracted intensive attention in recent years. One representative work is support vector machine (SVM). Although it works well in binary classification, SVM has a non-smooth loss function and can not naturally cover multi-class case. In this paper, we propose manifold regularized kernel logistic regression (KLR) for web image annotation. Compared to SVM, KLR has the following advantages: (1) the KLR has a smooth loss function; (2) the KLR produces an explicit estimate of the probability instead of class label; and (3) the KLR can naturally be generalized to the multi-class case. We carefully conduct experiments on MIR FLICKR dataset and demonstrate the effectiveness of manifold regularized kernel logistic regression for image annotation.

# 1. Introduction

Today, smart devices e.g. smart phone, table PC which equipped with a digital camera have become more and more popular, and people can easily produce millions or even billions of multimedia information such as personal photos or videos. However, it is not convenient to effectively manage the photos or videos at the semantic level, and therefore large scale image/video annotation through innovative machine learning methods has attracted intensive attention in recent years and been successfully deployed for many practical applications in multimedia, computer vision and image processing [14] [16] [17] 0.

There are a number of machine leaning algorithms have been employed for image annotation. One of the representative methods is support vector machine (SVM) that tries to find a separating hyperplane to maximize the margin between two classes.[12] SVM usually minimizes a hinge loss to train the maximum-margin classifier. Although hinge loss is a convex function, it is not differentiable and can not naturally be generalized to multi-class cases[15] .

On the other hand, it is a very expensive labor to label a large number of images to learn a robust model for image annotation. Then semi-supervised learning (SSL) has been employed for semi-automatic image annotation[7] [8] . SSL can improve the generalization ability with only a small number of labeled images by exploiting the intrinsic structure of all the training samples including labeled and unlabelled images[1] . The most traditional class of SSL methods is manifold regularization that tries to explore the geometry of intrinsic data probability distribution by penalizing the objective function along the potential manifold [1] [4] .

Considering the above analysis, in this paper, we replace hinge loss in SVM with logistic loss and propose manifold regularized kernel logistic regression (KLR) for web image annotation. Particularly, we employ the representative Laplacian graph to exploit the geometry of the underlying manifold. Compared to SVM, manifold regularized KLR has the following immediate advantages: (1) the KLR has a smooth loss function; (2) the KLR produces an explicit estimate of the probability instead of class label; (3) the KLR can naturally be generalized to the multi-class case; and (4) Laplacian regularization can well utilize the intrinsic structure of the data distribution. We carefully conduct extensive experiments on the MIR FLICKR dataset. The experimental results verify the effectiveness of Laplacian regularized KLR for web image annotation by comparison with the baseline algorithms.

The rest of this paper arranged as follows. Section 2 briefly reviews the related work of image classification. Section 3 presents the proposed manifold regularized KLR framework. Section 4 details the implementation of Laplacian regularized KLR. Section 5 demonstrates experimental results on the MIR FLICKR dataset. And Section 6 concludes the paper.

## 2. Related work

In recent years, there are many algorithms been proposed for multimedia retrieval including image annotation/classification, video indexing, and 3D object retrieval etc.

Briefly, the related image/video annotation methods can be divided into three categories based on the employed machine leaning schemes which are unsupervised, supervised, and semi-supervised learning.

Unsupervised learning methods use unsupervised machine learning methods such as nonnegative matrix factorization [3], clustering[10] to annotate images/videos.

Supervised leaning methods such as support vector machines [13], decision trees[11] aim to find the relationship between labels and visual features. Considering the growing large amount of samples, some active learning methods [14] are introduced to interactively select only effective samples for labeling.

Considering the heavy user labeling effort, semi-supervised learning methods exploit both a small number of labeled samples and a large number of unlabeled samples to boost the generalization of learning model and receive more and more intensive attention recently [9].

## 3. Manifold regularized kernel logistic regression

In semi-supervised image annotation, we are given a small number of labeled images $S_L = \{(x_i, y_i)\}_{i=1}^{l}$ and a large number of unlabeled images $S_U = \{x_j\}_{j=l+1}^{l+u}$, where $y_i \in \{+1, -1\}$ is the label of $x_i$ and $l, u$ denote the number of labeled and unlabeled images respectively. Typically, $l \ll u$. Under the assumption of semi-supervised learning, the labeled images $(x, y) \in S_L$ are drawn from a probability $P$, and unlabeled images $x \in S_U$ are simply drawn from the marginal distribution $P_X$ of $P$ where $P_X$ is a compact manifold $M$. This assumption indicates that the conditional distribution $P(y|x)$ varies smoothly along the geodesics in the underlying geometry of $M$ and then close images pairs induce similar conditional distribution pairs.

The manifold assumption is widely employed in SSL because it is a key point to precisely explore the local geometry of the potential manifold. Then the SSL problem can be written as the following optimization problem by incorporating an additional regularization term to exploit the intrinsic geometry:

$$min_{f \in H_k} \frac{1}{l}\sum_{i=1}^{l} \varphi(f, x_i, y_i) + \lambda_1 \|f\|_K^2 + \lambda_2 \|f\|_I^2 \qquad (1)$$

where $\varphi$ is a general loss function, $\|f\|_K^2$ penalizes the classifier complexity in an appropriate reproducing kernel Hilbert space (RKHS) $H_k$, $\|f\|_I^2$ is the manifold regularization term to penalize $f$ along the underlying manifold, and parameters $\lambda_1$ and $\lambda_2$ balance the loss function and regularization terms $\|f\|_K^2$ and $\|f\|_I^2$ respectively.

Although there are different choices for the manifold regularization terms $\|f\|_I^2$, Laplacian regularization is promising to preserve the local similarity. In this paper, we introduce the Laplacian regularized kernel logistic regression to web image annotation. In this paper, we employ logistic loss $\log(1 + e^{-f})$ for the loss function to construct a kernel logistic regression (KLR) model. Logistic loss is equivalent to the cross entropy loss function. Some traditional loss functions are plotted in Figure 1. The dashdot line is 0-1 loss, the dotted line is Hinge loss, and the solid line is logistic loss. From Figure 1 we can see that the negative log-likelihood loss is smooth and has a similar shape to Hinge loss that used for the SVM. Hence it is expected that the KLR has similar performance with the SVM.

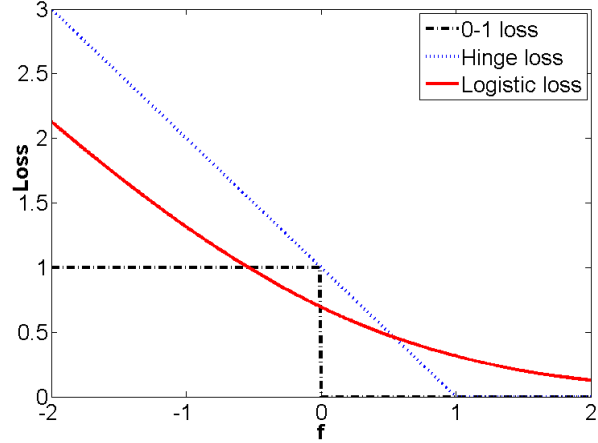

Figure 1. several loss functions

Therefore, we incorporate Laplacian regularized term into the objective function with logistic loss. And then we have the following equivalent optimization problem.

$$min_{f \in H_k} -\frac{1}{l}\sum_{i=1}^{l} \left(y_i \log \frac{1}{1+e^{-f(x_i)}} + (1-y_i) \log \left(1 - \frac{1}{1+e^{-f(x_i)}}\right)\right) + \lambda_1 \|f\|_K^2 + \lambda_2 \mathbf{f}^T L \mathbf{f} \quad (2)$$

where $\mathbf{f} = [f(x_1), f(x_2), \cdots, f(x_{l+u})]^T$, $L$ is the graph Laplacian given by $L = D - W$. Here $D$ is a diagonal matrix given by $D_{ii} = \sum_{j=1}^{l+u} W_{ij}$ where $W$ is the edge weight matrix for data adjacency graph.

**Theorem 1:** The minimization of (2) w.r.t. $f \in H_k$ exits and has the following representation

$$f^* = \sum_{i=1}^{l+u} \theta_i K(x_i, x). \quad (3)$$

where $K(x_i, x)$ is a valid kernel in RKHS.

The representer theorem shows the solution of (2) exists and has the general form in terms of the expansion of both labeled and unlabeled images. The proof of this representer theorem can be sketched as below.

**Proof:** Suppose the subspace $S = span\{K(x, x_i) | 1 \leq i \leq l + u\}$ is spanned by the kernels centered at labeled and unlabeled images and $S^\perp$ is the orthogonal complement of $S$. Thus any $f \in H_k$ can be represented as $f = f_S + f_{S^\perp}$, wherein $f_S$ is the projection of $f$ onto $S$ and $f_{S^\perp}$ is the projection of $f$ onto $S^\perp$. Then we have $\|f\|^2 = \|f_S\|^2 + \|f_{S^\perp}\|^2 \geq \|f_S\|^2$.

On the other hand, $K$ is a valid (symmetric, positive definite) kernel in RKHS and graph Laplacian $L$ is semi-definite positive. Thus $G(\|f\|) = \lambda_1 \|f\|_K^2 + \lambda_2 \|f\|_I^2 = \lambda_1 \mathbf{f}^T K \mathbf{f} + \lambda_2 \mathbf{f}^T L \mathbf{f}$ is a monotonically increasing real-valued function on $\|f\|$. Then we have $G(\|f\|^2) \geq G(\|f_S\|^2)$. This implies that $G(\|f\|)$ is minimized if $f$ lies in the subspace $S$.

Note the reproducing property of the kernel $K$, then $f(x_i) = \langle f, K(x_i, x) \rangle = \langle f_S, K(x_i, x) \rangle + \langle f_{S^\perp}, K(x_i, x) \rangle = f_S(x_i)$. Therefore, the solution of the optimization problem (2) can be obtained when $f$ lies in the subspace $S$, that is $f^* = \sum_{i=1}^{l+u} \theta_i K(x_i, x)$. This completes the proof of Theorem 1. ∎

Substituting (3) into (2), we have the following Laplacian regularized kernel logistic regression

$$min_{\theta \in R^{l+u}} -\frac{1}{l}\sum_{i=1}^{l}\left(y_i \log \frac{1}{1+e^{-K(x_i,x)\theta}} + (1-y_i) \log \left(1 - \frac{1}{1+e^{-K(x_i,x)\theta}}\right)\right) + \lambda_1 \theta^T K \theta + \lambda_2 \theta^T KLK\theta. \quad (4)$$

Because the objective function is differential, we have many iterate numerical solutions for problem (4), e.g. gradient descent algorithm, Newton-Raphson method. In the next section, we describe the conjugate gradient algorithm employed in this paper to solve problem (4).

## 4. Algorithm

In this section, we employ the conjugate gradient algorithm to optimize problem (4).

The gradient of the objective function in (4) can be written as:

$$\nabla f(\theta) = -\frac{\log e}{l} \sum_{i=1}^{l} \left( y_i K(x_i, x) e^{-K(x_i,x)\theta} \frac{1}{1+e^{-K(x_i,x)\theta}} + (1-y_i)\left(-K(x_i,x)\frac{1}{1+e^{-K(x_i,x)\theta}}\right) \right)$$

$$+ \lambda_1 (K + K^T)\theta + \lambda_2 (KLK + (KLK)^T)\theta$$

Then we have the optimization procedure of conjugate gradient algorithm as below:

Step 1: Initialize $\theta^0, \beta, d^0 = -\nabla f(\theta^0), 0 < \varepsilon \ll 1, k = 0$.

Step 2: Do

$$\theta^{k+1} = \theta^k + \beta d^k,$$

$$d^{k+1} = -\nabla f(\theta^{k+1}) + \frac{\|\nabla f(\theta^{k+1})\|^2}{\|\nabla f(\theta^k)\|^2} d^k.$$

Until $|f(\theta^{k+1}) - f(\theta^k)| < \varepsilon$.

Step 3: $\theta^* = \theta^k$.

The optimization of problem (4) is efficient and effective due to the smoothness character of the objective function. From the illustration of different loss functions in Figure 1, the logistic loss can achieve almost equivalent performance to hinge loss. In the following section we describe the comparison experiments.

## 5. Experiments

To evaluate the effectiveness of the proposed algorithm, we carefully conduct web image annotation experiments on the MIR Flickr dataset [6] that is offered by the LIACS Medialab at Leiden University, the Netherlands and introduced by the ACM MIR Committee in 2008 as an ACM sponsored image retrieval evaluation. The dataset contains 25,000 images of 38 categories including animals, baby, baby*, bird, bird*, car, car*, clouds, clouds*, dog, dog*, female, female*, flower, flower*, food, indoor, lake, male, male*, night, night*, people, people*, plant_life, portrait, portrait*, river, river*, sea, sea*, sky, structures, sunset, transport, tree, tree*, water. Figure 2 shows some example images in the dataset.

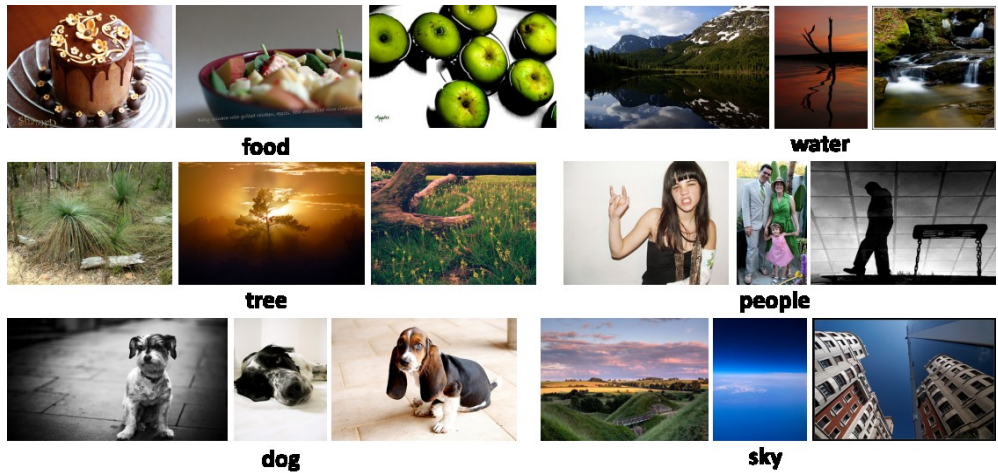

Figure 2. Example images of the MIR Flickr dataset

In our experiments, 25000 images are randomly split into equal-sized two parts as training set and test set. And for the semi-supervised learning experiments, we randomly select the same number $l/2 = \{5, 10, 20, 50, 100\}$ for positive and negative labeled samples for each class and all the rest samples are unlabeled ones.

In our experiment, we employ GIST descriptor extracted by Guillaumin [5] . GIST descriptor is a biologically-inspired image feature which describes image features from the visual cortex cognitive mechanism.

We compare the proposed Laplacian kernel logistic regression algorithm with some baseline algorithms including SVM classifier and kernel logistic regression method. For Laplacian kernel logistic regression method, parameter $\lambda_1$ and $\lambda_2$ are tuned from the candidate set $\{10^e | e = -4, -3, -2, -1\}$.

In our experiments, we measure the performance by using the average precision

(AP) and mean average precision (mAP). Particularly, AP and mAP are computed by using the PASCAL VOC method [2].

$$AP = \frac{1}{11}\sum_t \left[\max_{k \geq t} p(k)\right], t \in \{0, 0.1, 0.2, \cdots, 1.0\}$$

and

$$mAP = \frac{\sum_{i=1}^{\#} AP_i}{\#\{visual\ object\ classes\}}$$

Where $p(k)$ is the precision at recall $k$.

Figure 3 shows the average precision (AP) performance of some representative objects. Each subfigure of this figure shows the performance curves of a particular category from sky, sunset, structures, clouds, clouds*, animals, indoor, people*, tree, female, female*, male, transport, water. The x-coordinate of each subfigure is the number of the labeled (unlabelled) samples in the training set and the y-coordinate is the average precision. It shows that kernel logistic regression can achieve similar performance to SVM classifier and Laplacian kernel logistic regression outperforms the baselines in most cases.

Figure 4 illustrates the mean average precision (mAP) boxplots of different methods. There are five subfigures each of which demonstrates the performance of a particular number of labeled and unlabeled samples. The mAP performance also shows that Laplacian kernel logistic regression algorithm performs better than baseline methods.

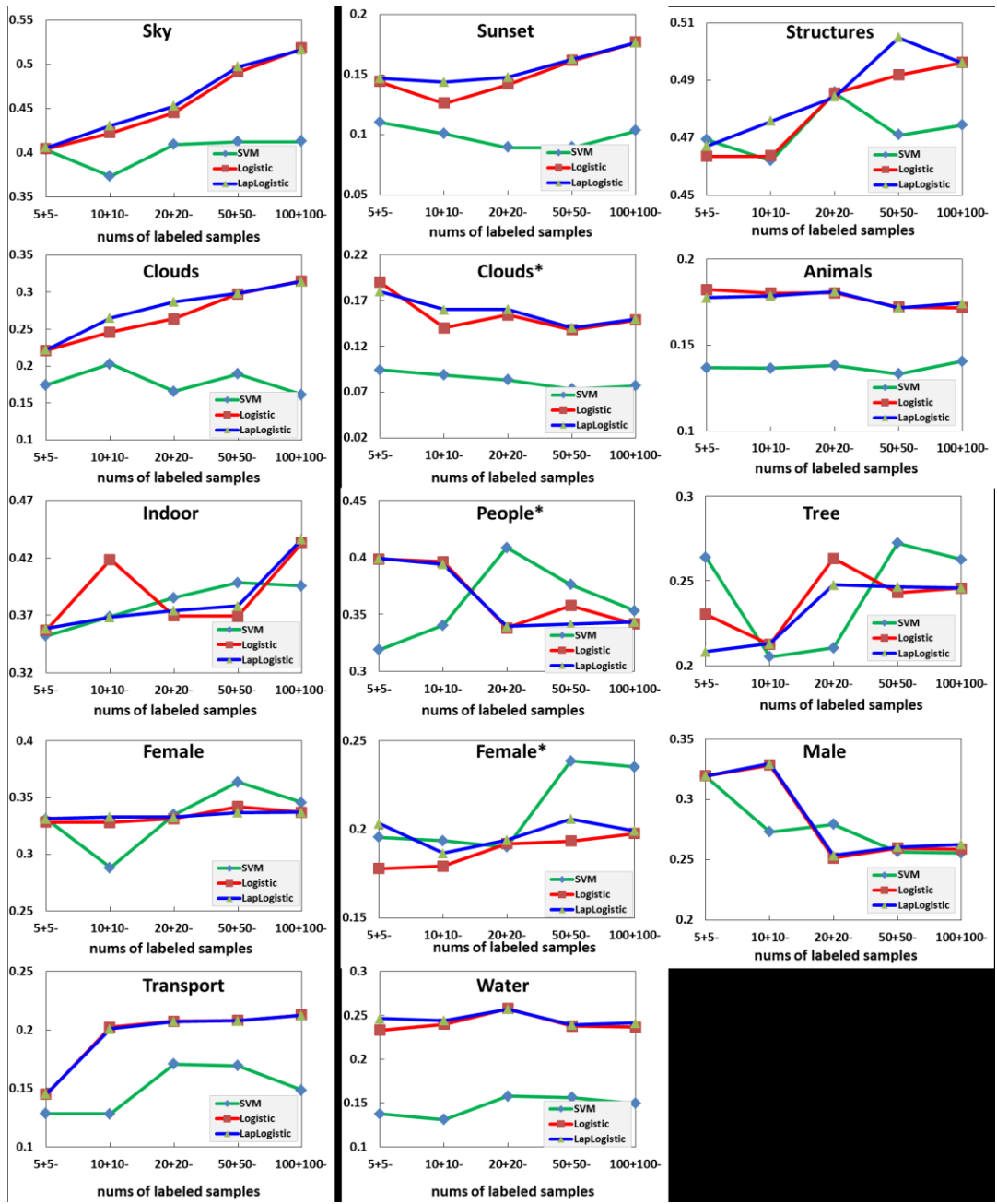

Figure 3. Average precision of several representative objects.

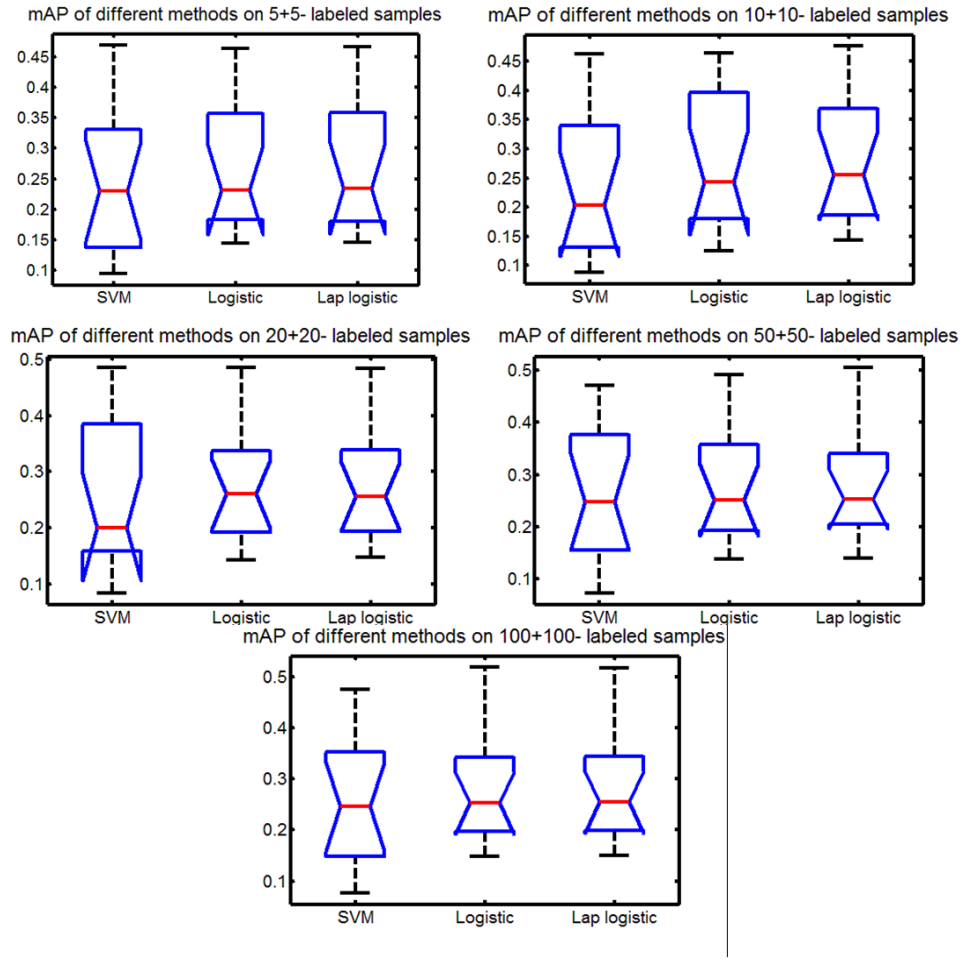

Figure 4. The mAP performance of different methods.

## 6. Conclusion

This paper studies manifold regularized kernel logistic regression (KLR) for web image annotation. Technically, we develop Laplacian regularized kernel logistic regression and implement image annotation task on MIR Flickr dataset. Compared to the representative SVM classifier, the KLR has a smooth loss function and produces an explicit estimate of the probability instead of class label. The carefully conducted experiments demonstrate the effectiveness of manifold regularized kernel logistic regression for image annotation.

In the future, we will apply the proposed Laplacian regularized kernel logistic regression to other applications. We will also further extend the proposed method to other manifold regularizations and explore the relation of the different regularizations.